\documentclass[conference]{IEEEtran}
\IEEEoverridecommandlockouts
\usepackage{cite}
\usepackage{amsmath,amssymb,amsfonts}
\usepackage{algorithmic}
\usepackage{float}
\usepackage{graphicx}
\usepackage{textcomp}
\usepackage{booktabs}
\usepackage{xcolor}
\def\BibTeX{{\rm B\kern-.05em{\sc i\kern-.025em b}\kern-.08em
    T\kern-.1667em\lower.7ex\hbox{E}\kern-.125emX}}
\begin{document}

\title{Global License Plate Dataset\\
}

\author{\IEEEauthorblockN{Siddharth Agrawal}
\IEEEauthorblockA{\textit{Ahmedabad University} \\
siddharth.a@ahduni.edu.in}
}

\maketitle

\begin{abstract}
In the pursuit of advancing the state-of-the-art (SOTA) in road safety, traffic monitoring, surveillance, and logistics automation, we introduce the Global License Plate Dataset (GLPD). The dataset consists of over 5 million images, including diverse samples captured from 74 countries with meticulous annotations, including license plate characters, license plate segmentation masks, license plate corner vertices, as well as vehicle make, colour, and model. We also include annotated data on more classes, such as pedestrians, vehicles, roads, etc. We include a statistical analysis of the dataset, and provide baseline efficient and accurate models. The GLPD aims to be the primary benchmark dataset for model development and finetuning for license plate recognition.
\end{abstract}

\begin{IEEEkeywords}
License Plate Recognition, License Plate Detection, Computer Vision, Deep Learning, Machine Learning
\end{IEEEkeywords}

\section{Introduction}
Automatic License Plate Recognition is an essential task for several applications related to Intelligence Transport Systems, and plays a role in traffic monitoring, surveillance, logistics, and other related applications. It requires robust performance in complex and challenging environments. Accuracy, model size, and efficiency, are key metrics for its application in the real-world and deployment at scale. 

At present, intelligent transportation systems (ITS) have been widely applied in various fields, e.g., improving transportation security and enhancing productivity. \cite{anagnostopoulos_anagnostopoulos_loumos_kayafas_2006a}.

\subsection{Related Work}
\subsection{License Plate Recognition}
License Plate Recognition (LPR) has been widely studied in the fields of Computer Vision and Machine Learning.

\textbf{Character Object Detection Methods:}
traditional License Plate Recognition (LPR) methods typically comprise of two stages: explicit character detection followed by independent character recognition. Examples of heuristic algorithms used for character detection include Maximally Stable Extremal Regions (MSER) \cite{hsu_chen_chung_2013a}, Connected Component Analysis (CCA) \cite{anagnostopoulos_anagnostopoulos_loumos_kayafas_2006a}.

\begin{figure}[H]
    \centering
    \includegraphics[width=\linewidth]{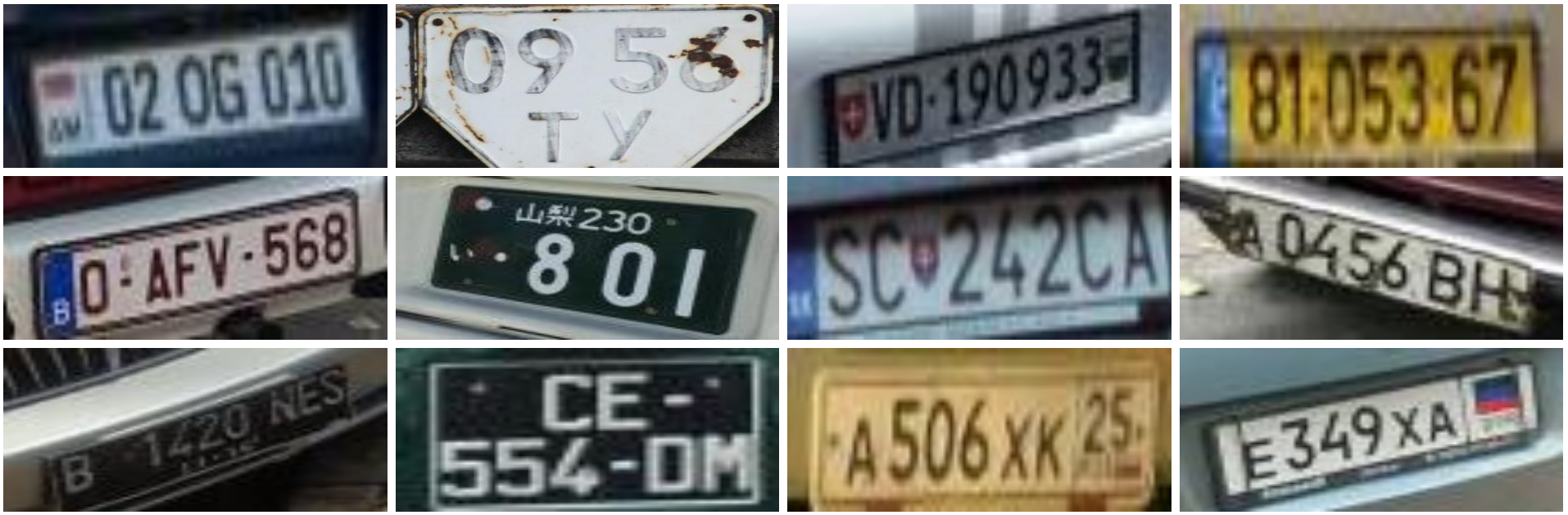}
    \caption{License Plate Samples from the Dataset}
    \label{fig:enter-label}
\end{figure}

\textbf{Sequential Methods:} 
with the advent of Deep Learning, Sequential methods have been developed to address this issue. Typically, a sliding window is used to densely extract character probabilities or features from the plate images, and a sequential model is then employed for character recognition. CNNs have been used for feature extraction \cite{li_shen_2016a}, while Sweep OCR \cite{bulan_kozitsky_ramesh_shreve_2017} can be used to obtain character probability sequences. Recurrent Neural Networks (RNN) with Long Short-Term Memory (LSTM) \cite{li_shen_2016a}.

\textbf{Multi-classifier-Based Methods:}
 To avoid extracting the position information, Multi-classifier based methods have been proposed \cite{xu_yang_meng_lu_huang_ying_huang_2018a,spanhel_sochor_juranek_herout_marsik_zemcik_2017,deep_automatic_license_plate_recognition_system__proceedings_of_the_tenth_indian_conference_on_computer_vision,goncalves_diniz_laroca_menotti_robson_schwartz}. In these methods, global features are extracted from the entire plate image, and then fed into different classifiers which are expected to automatically focus on different character regions. However, the classifiers primarily designed for recognition are difficult to accurately localize the characters and the character distributions at different positions are highly inconsistent \cite{xu_yang_meng_lu_huang_ying_huang_2018a}. 

\textbf{Segmentation-based methods}: Alternatively, Segmentation-based methods formulate LPR into a semantic segmentation task, producing recognition results over pixels rather than one semantic label for each character. Such methods decompose the LPR task into a segmentation-based task, wherein the characters are annotated via polygons or bounding boxes, and the task becomes to segment the pixels or predict bounding boxes with characters as labels, for each character in the license plate image. \cite{zhang_wang_zhuang_2021}

\subsubsection{Existing Datasets}

\textbf{CCPD dataset:} \cite{xu2018towards} dataset of Chinese License Plates. Contains 290K images total. 100K for training, 100K for validation, and 100K for test. We use this official split appropriately, using the 100K images in training set for training, 100K images in validation set for selection of model, and hyperparameters, and the last set of 100K test images is reserved solely for testing performance at the end of research/experimentation/once the final model was finalised.

\textbf{AOLP dataset:} \cite{hsu2012application} dataset of Taiwanese license plates. Contains 1874 images with alphanumeric characters, devided into 3 subsets: access control (AC), law enforcement (LE), and road patrol (RP), with 681, 582, and 611 images respectively. The dataset's license plate bounding box annotations had to be reannotated due to the poor quality of the original annotations. 

\textbf{MediaLab dataset:} \cite{media_lab} dataset of Greek license plates. Contains 706 images with alphanumeric characters, and does not provide an official train/validation split. 

\textbf{Iraqi License Plate datasets:} Iraqi License Plates form a unique challenge due to the use of arabic symbols, and even in our dataset, we could only manage to collect 400 samples. We hope to include more license plates in the future or combine existing datasets but information regarding consent and license is often vague. Some are available on Kaggle such as \cite{iraqi_license_plates} that has a compilation of 1052 images.

\textbf{Indian License Plate datasets:} Some small scale Indian license plate datasets exist such as  \cite{tanwar_tiwari_chowdhry_2021}, \cite{narenbabu_sowmya_soman_2019}, \cite{Indian_Truck}. However, they are again small in scale, ranging from a few hundred to 5000 license plates.

\subsubsection{Scene Text Recognition}
Scene text recognition has been a research interest in the field of computer vision. It is currently using an encoder decoder framework with attention mechanism, to learn the mapping between input images and output sequences in a data-driven manner. License plate recognition can be seen as a special case of general scene text recognition tasks, for they share the similarity of continuous text. A flexible Thin Plate Spline transformation was proposed by \cite{shi_yang_wang_lyu_yao_bai_2019} for processing a variety of text irregularities. \cite{luo_jin_sun_2019} proposed a multi-object rectifier attentional network (MORAN) for general scene text recognition. MORAN consists of a multi-object rectifier network and an attention-based sequence recognition network, which can read both regular and irregular scene texts. The network reduces the difficulty of recognition and makes it easier for the attention-based sequence recognition network to read irregular text. \cite{li_wang_shen_zhang_2018} achieved state-of-the-art performance on both regular and irregular scene text recognition with the encoder-decoder framework in LSTM and a two-dimensional attention module. This research is similar to ours, but in our framework, the Bi-LSTM and one-dimensional attention modules are adopted to accurately locate character features and enhance the recognition performance of the model. 

PARSeq came up with the current state-of-the-art in STR. Furthermore, with recent developments in hardware and GPUs, it enables shallower networks with high parallelism such as transformers to produce faster and more accurate results, leading to higher accuracy and faster inference times, which makes it ideal for a task such as LPR, where inference speed has to be maximised. PARSeq iteratively refines the predictions using an Ensemble of Language Modelling models, wherein, it trains using an ensemble of language modelling methods with shared weights, and in inference, can either decode parallelly or autoregressively. The predictions can be further refined using iterative refinement. ParSeq not only has the highest accuracy in the STR task, being the current state-of-the-art, but also has some of the lowest inference times on capable GPUs: ~12 milliseconds for non-autoregressive decoding, and ~15 milliseconds for autoregressive decoding on an NVIDIA Tesla A100 GPU. 

\subsection{Motivation}

While there are many available license plates from various countries, open-access data availability for many countries remains low. Even for the countries where data is available, the datasets are small in scale, with the largest research dataset: CCPD \cite{xu2018towards} having only 350k images. Moreover, most existing datasets have saturated in accuracy with state-of-the-art models achieving more than 99.5\% accuracy \cite{scr_net} due to the relatively lower difficulty of the license plate detection and recognition (LPDR) tasks, and validation leaks of the same license plate texts as described in \cite{near-duplicates}, and the relatively low diversity of datasets. They largely focus on samples that are clearer and free of distortions. These do not have a diversity of fonts, are either small-scale or not publicly available, and are not focused on challenging scenarios that reflect real-world conditions, such as varying lighting conditions, partial occlusions, and diverse environmental backgrounds. While CCPD addresses some of these issues, it is still relatively smaller in scale and does not provide variation in license plate designs, country, or fonts. The limitations of existing datasets underscore the need for a more extensive, diverse, and challenging license plate dataset, which our dataset aims to address. By curating a larger, more diverse collection of license plates with a focus on challenging scenarios, we aim to provide researchers with a resource that better aligns with the complexities encountered in real-world license plate recognition tasks. 

Furthermore, fine-tuning for license plate recognition in target countries is often crucial, and existing models trained on limited datasets struggle to generalize effectively across diverse countries. The inadequacy of existing models in achieving broad generalization stems from the fact that many datasets predominantly focus on specific regions, resulting in a lack of diversity and representative samples from various countries. As a consequence, the models trained on these datasets may lack the adaptability required to recognize license plates across a global spectrum. Additionally, variations in license plate formats, fonts, and design elements further amplify the challenge. Our approach addresses this limitation by assembling a dataset that spans multiple countries, encompassing a wide range of license plate variations encountered globally. This dataset not only facilitates the development of more robust and universally applicable models but also caters to the need for fine-tuning specific to individual countries, ultimately enhancing the adaptability and performance of license plate recognition systems on a global scale.

Many papers such as \cite{baek_matsui_aizawa_2021}, \cite{loginov_2021}, and \cite{bautista_atienza_2022a}, outline the increased efficacy of real-world STR datasets as compared to synthetic datasets. Thus, we make a similar claim that synthetic datasets of license plates alone would not generalise well to real-world license plates. Thus, arising the need for such a dataset with samples from the real world.

\subsection{Novel Dataset Challenges:}

License plate detection and recognition are integral components in numerous real-world applications, including traffic law enforcement, surveillance systems, toll booth systems, logistics, and car parking registration management. While rotations, illuminations, and shear-based distortions are common and well-studied challenges, our proposed global license plates dataset introduces additional complexities, including but not limited to:

\begin{enumerate}
   \item \textbf{ Varied Fonts and Formats:} The dataset encompasses diverse fonts and formats of license plates encountered across different regions and states, contributing to the challenge of recognition. Global license plates exhibit a wide variety of fonts and formats, posing a challenge for recognition models that are often trained on datasets with limited font and format diversity. 
    \item \textbf{Multilingual Text: }License plates of some countries feature multilingual text, reflecting the linguistic diversity of texts. This adds a layer of complexity for recognition models not trained on such diverse linguistic patterns. The presence of license plate text in multiple languages further complicates recognition tasks, requiring models to be adept at handling linguistic diversity.
\begin{figure}[H]
        \centering
        \includegraphics[width=1\linewidth]{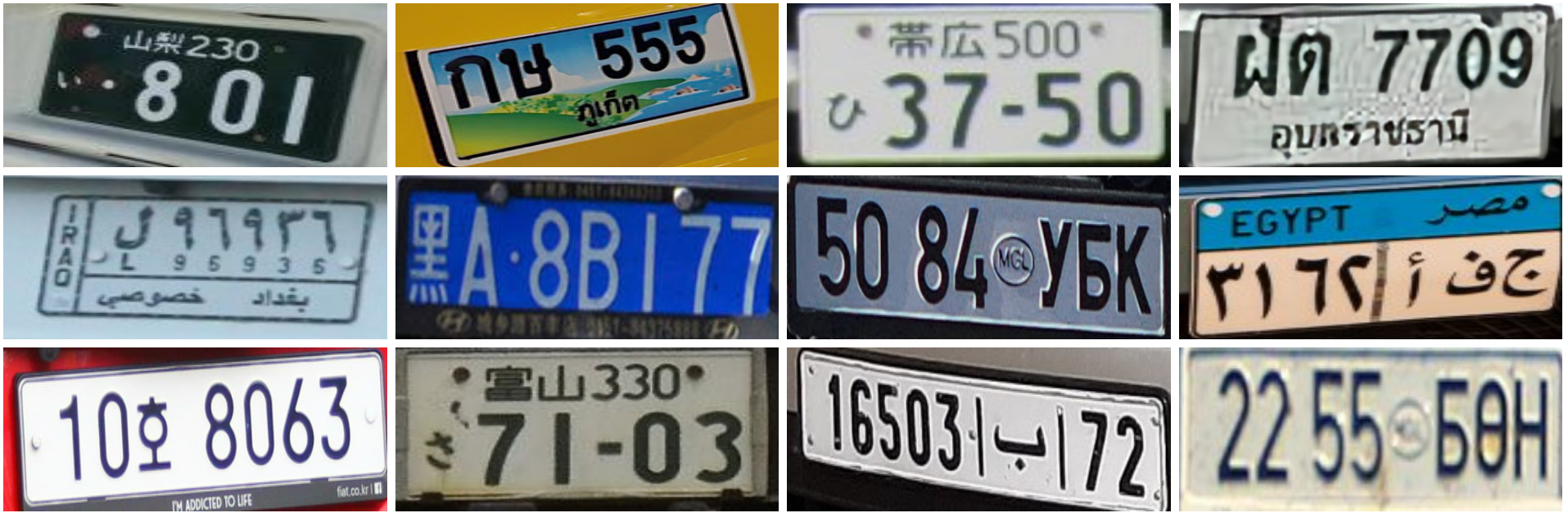}
        \caption{Varied Languages and Scripts}
        \label{fig:enter-label}
    \end{figure}
    \item \textbf{Dynamic Environmental Conditions:} Real-world scenarios involve dynamic changes in environmental conditions, such as varying light intensities, weather conditions, and background complexities, making license plate recognition more challenging.
    \item \textbf{Crowded and Cluttered Scenes: }The dataset includes images captured in crowded and cluttered scenes, simulating scenarios encountered in urban areas or busy traffic, where license plates may be partially obscured or surrounded by visual noise.
    \item \textbf{Extreme scales:} The dataset also captures extreme variations in scales; ranging from only license plates in the entire image (with no vehicle) to extremely small license plates of faraway vehicles.
   \item \textbf{Non-Standardized Plate Mounting:} License plates can exhibit non-standardized mounting, including variations in positioning: mounting on the top of the vehicle or inside the vehicle behind the windshield instead of on the bonnet, further increasing the difficulty of accurate detection and recognition.
\begin{figure}[H]
    \centering
    \includegraphics[width=\linewidth]{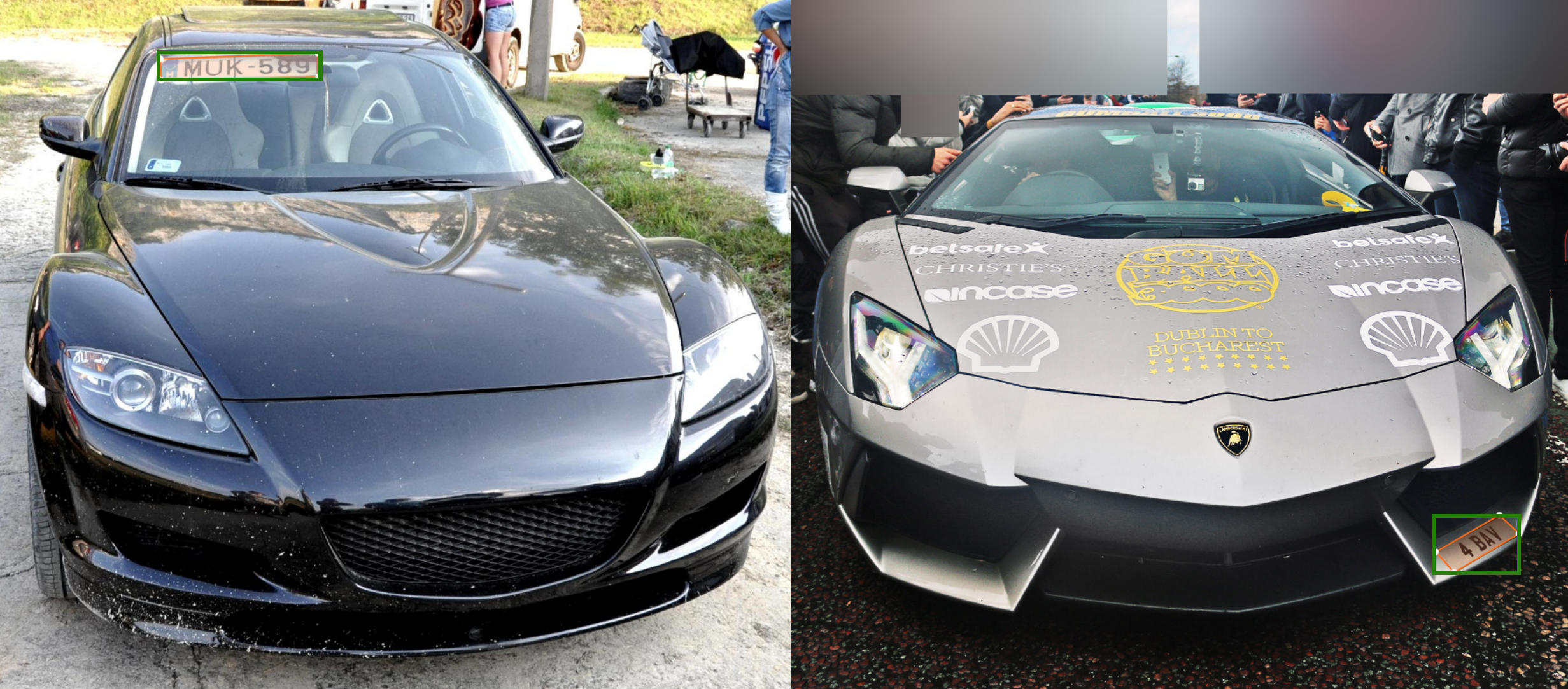}
    \caption{Examples of Non-Standardized Plate Mounting}
    \label{fig:enter-label}
\end{figure}
  \item \textbf{Varying Plate Designs:} Different countries employ unique designs and graphical elements on their license plates, necessitating a recognition system capable of accommodating diverse visual characteristics.
  \item \textbf{Environmental Contexts:} License plates are captured in various environmental contexts, including urban and rural settings, varying lighting conditions, and diverse weather conditions, demanding robust performance across different scenarios.
   \item \textbf{Spatial and Geographical Challenges:} The dataset includes plates from countries with distinct spatial and geographical features, such as densely populated urban areas, remote rural regions, and challenging terrains, adding complexity to recognition across diverse landscapes.
\end{enumerate}

 The Global License Plate Dataset (GLPD) encapsulates a rich variety of vehicles and license plate scenarios, offering a diverse and comprehensive resource for advancing research in Deep Learning and Computer Vision. The dataset showcases an extensive spectrum of vehicles, each encapsulating distinctive characteristics and scenarios, contributing to the complexity of real-world license plate recognition tasks.

 \section{Methodology}
\subsection{Data Collection}

The data in this paper is collected from several open-access datasets.

\textbf{Platesmania.com:} the majority of the data was collected from Platesmania.com. The site has a collection of many license plate images from 74 countries. PlatesMania is a website dedicated to license plates. It allows users to share and explore information about license plates from around the world. Users can upload photos of license plates, share details about them, and discuss various topics related to license plates. The users add text labels for the license plates at the time of uploading the images. As mentioned on their site, copying of data is allowed as long as an obligatory link is provided. The dataset is not only very varied in terms of number of countries, but also contains various vehicle types such as motorcycles, tractors, trucks, race cars, busses, cars, etc. Thus, this dataset is a large untapped resource for data availability for License Plate Detection and Recognition (LPDR) tasks. The GLPD is meticulously curated to encompass an extensive variety of vehicles and license plate scenarios, reflecting diverse real-world situations for robust training and evaluation. Encompassing a spectrum of vehicle types, from abandoned and damaged vehicles to specialized units like aerial device trucks and concrete mixers, the dataset captures nuances such as improper parking, road accidents, and traffic code violations. Notably, it includes scenarios featuring deliberately changed license plates, non-standard plates, and instances of vinyl wrapping. Covering vehicles such as electric, armored, and fire appliances, as well as unconventional categories like spyspots, tractors, ice-cream trucks, aerial vehicles, etc. The GLPD ensures a comprehensive representation of license plate scenarios. This diverse compilation is pivotal for advancing the capabilities of machine learning models to adeptly handle a broad array of real-world challenges in license plate recognition.

\subsection{Dataset Labeling and Annotations}

\begin{figure}[H]
    \centering
    \includegraphics[width=0.5\linewidth]{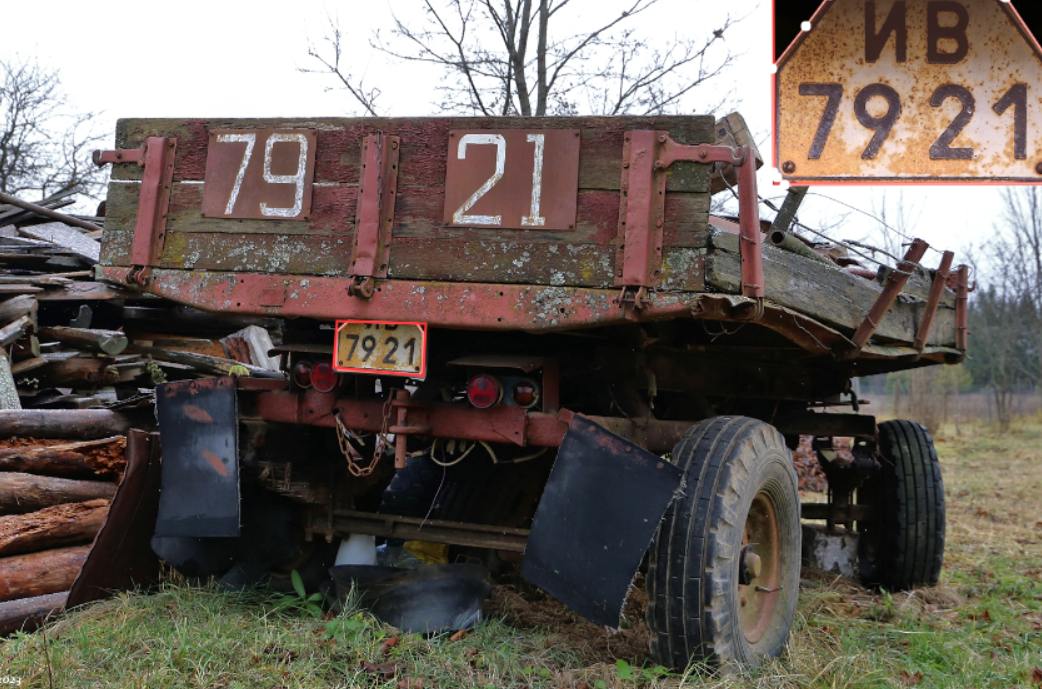}
    \label{fig:enter-label}
    \centering
    \includegraphics[width=0.5\linewidth]{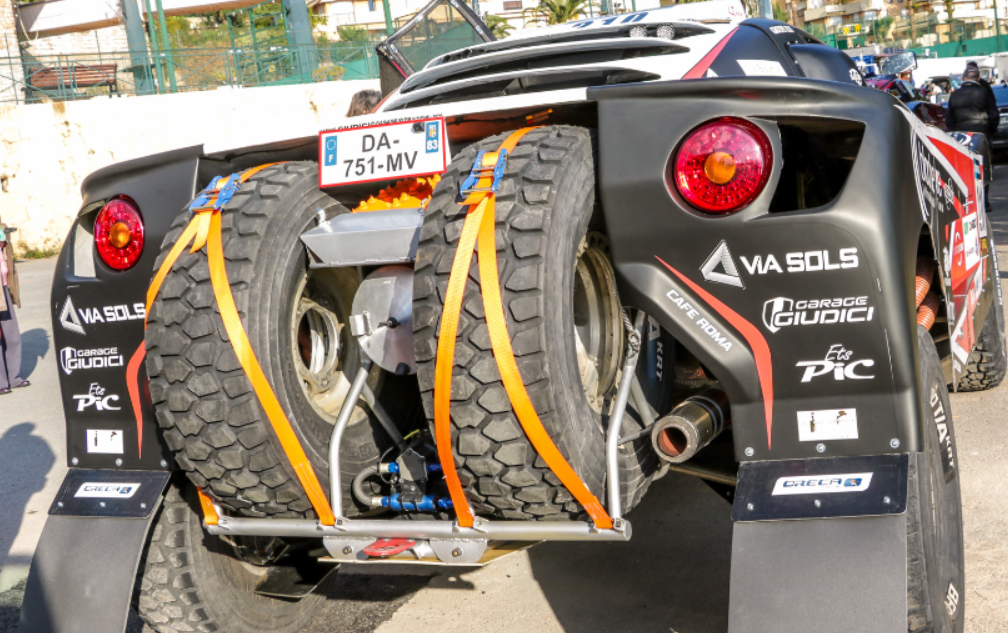}
    \caption{Example Annotations}
    \label{fig:enter-label}
\end{figure}

The license plates bounding box annotations were generated through an ensemble of large semi-supervised object detection models trained using \cite{efficient-teacher}. All of the labels were then manually verified extensively several times, assuring for correct label position, correct localisation of four corners, and matching with license plate text as seen on Platesmania. We used clean lab confidence learning \cite{confident_learning_cleanlab} to identify annotations likely to be erroneous and clean the dataset faster manually.

\begin{figure}[H]
    \centering
    \includegraphics[width=1\linewidth]{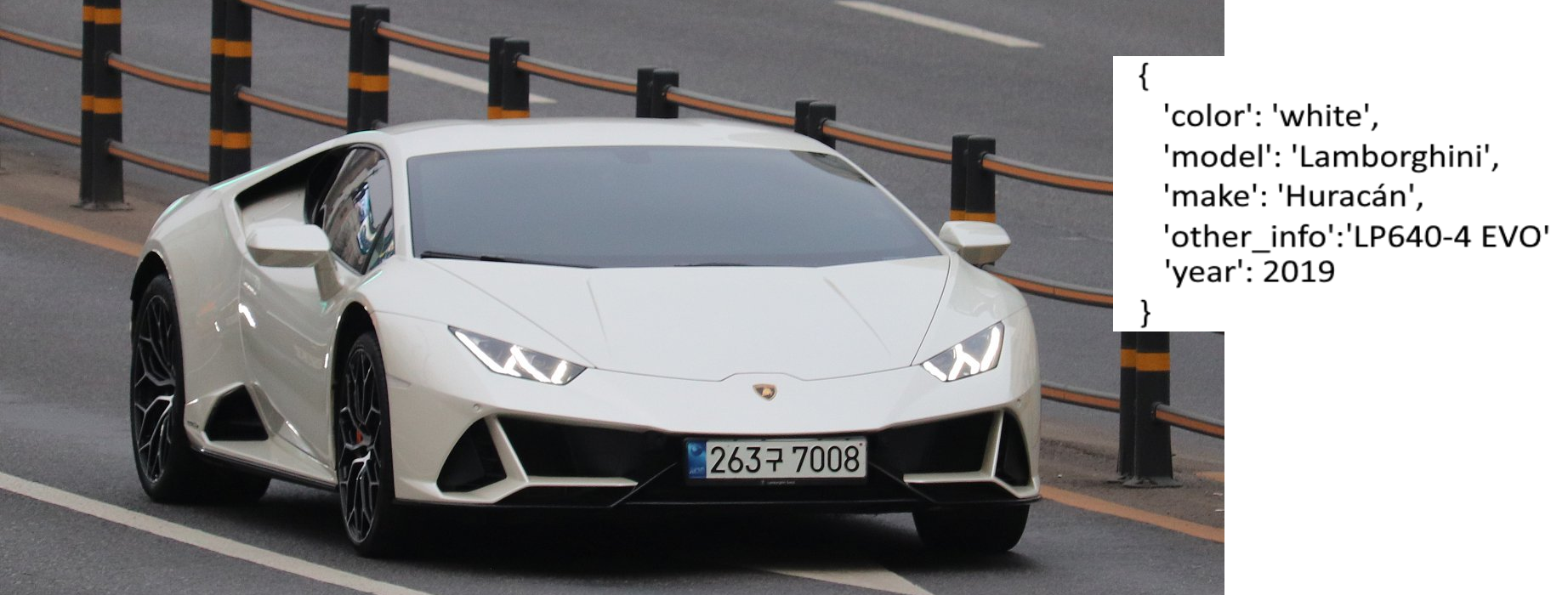}
    \caption{Example of Labels: Car make, model, year, color, and other information}
    \label{fig:enter-label}
\end{figure}

The dataset comprises comprehensive annotations, including the precise location of the four corners of the license plates, the polygon points outlining the instance segmentation of the license plate, and the bounding box information. The dataset includes valuable information the annotation for four vertices provides the exact (x, y) coordinates, accurately representing the borders of the license plate for robust object segmentation. We also quantify license plate blurriness, quantified through Variance of the Laplacian transform of the Image.  

Considering real-world scenarios where license plates may exhibit non-planar characteristics, such as bending, curvature, or being printed on the surface of a vehicle, we provide segmentation maps for such instances. This is particularly relevant when the license plate does not form a proper quadrilateral, ensuring that our dataset accommodates diverse and challenging scenarios commonly encountered in practical applications. 

\begin{figure}
    \centering
    \includegraphics[width=\linewidth]{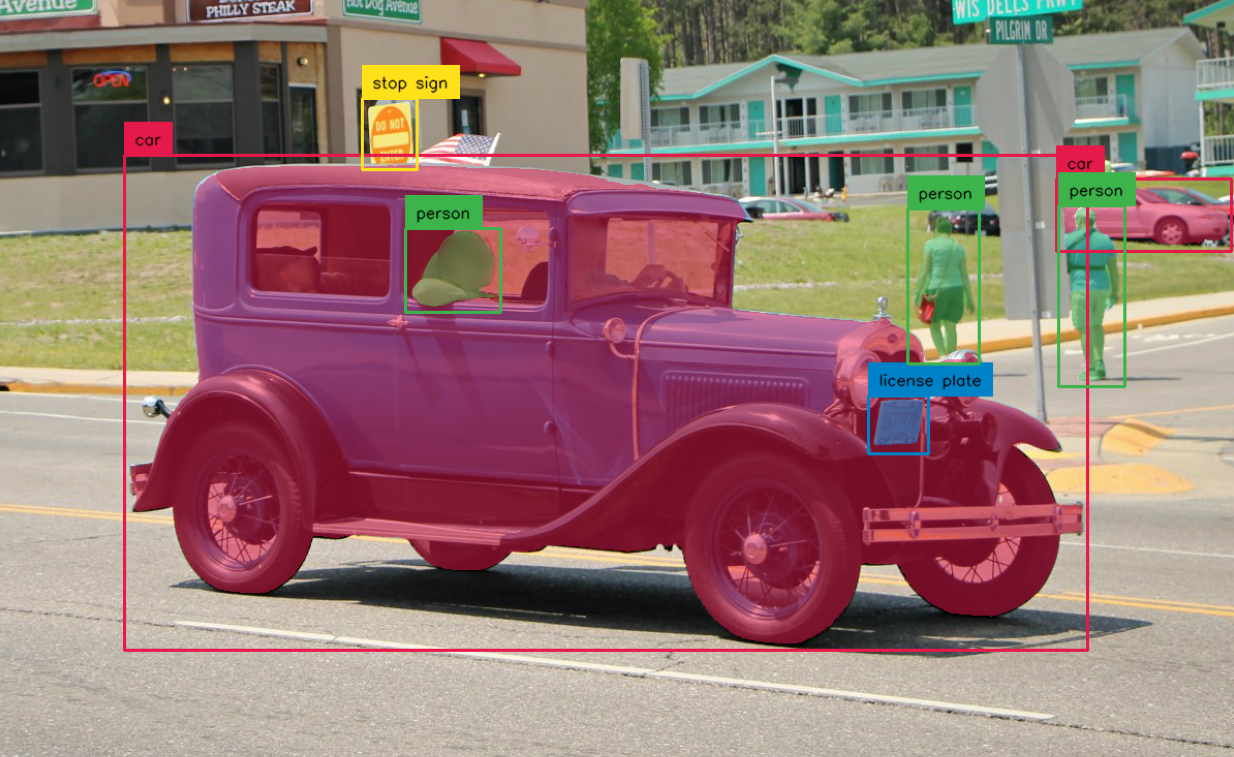}
    \caption{Example of multi-class Annotation labels: Bounding Boxes and Instance Segmentations}
    \label{fig:enter-label}
\end{figure}
Additionally, for approximately 20\% of the dataset, we have incorporated all COCO categories, complete with instance segmentation annotations and bounding boxes, further enhancing the dataset's versatility. The dataset also provides information about vehicle model, make, and year where possible.

Moreover, we incorporated the brightness and contrast information of the images, recognizing their significance in varying lighting conditions. These annotations contribute to a richer dataset, empowering researchers in Computer Vision and Deep Learning to explore and develop models that are not only accurate in typical scenarios but also resilient in the face of complex real-world challenges.

\begin{figure*}
    \centering
    \includegraphics[width=1\linewidth]{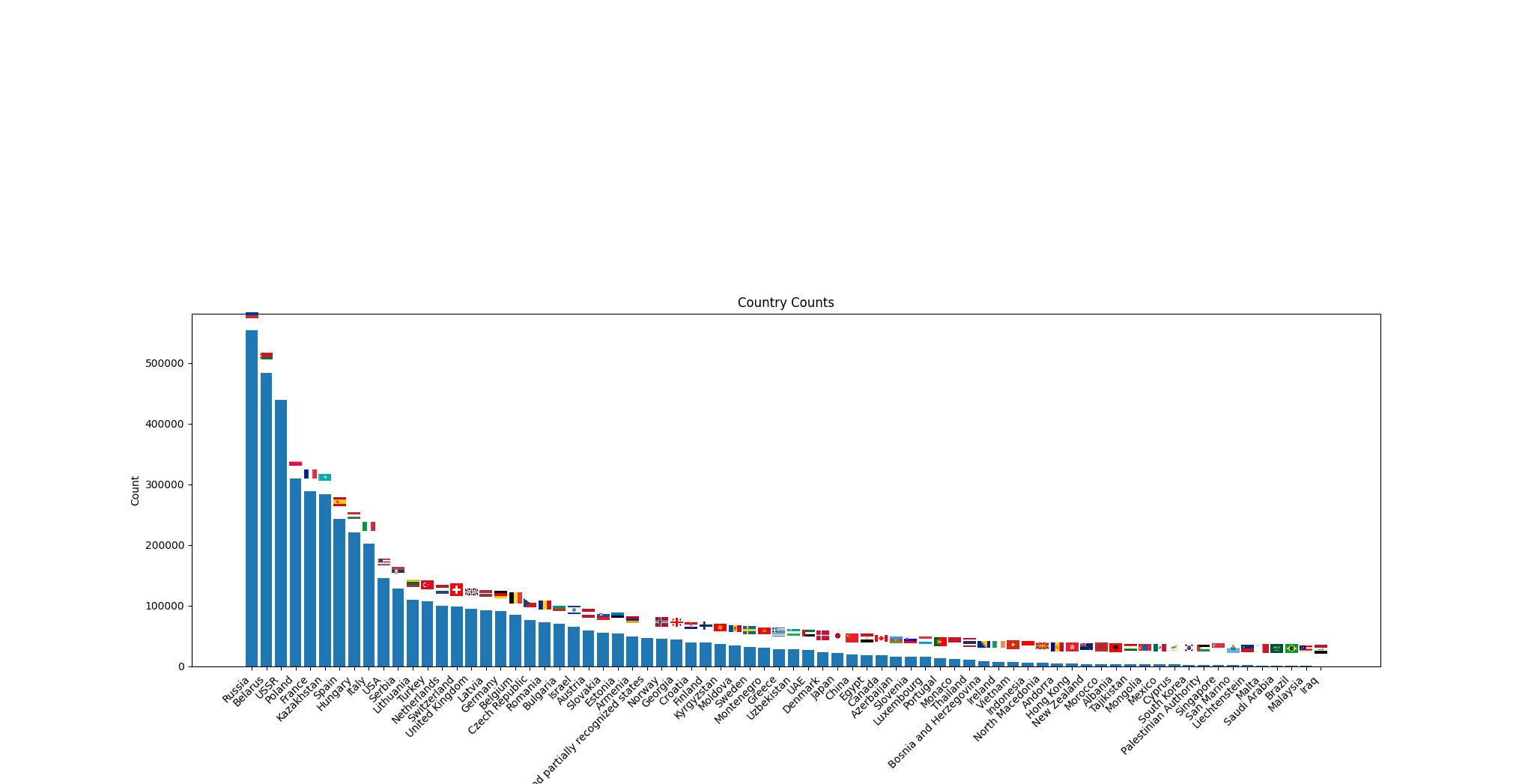}
    \caption{The Number of License Plate Images for each Country in the Dataset}
    \label{fig:countr_hist}
\end{figure*}

\subsection{Dataset Splits}

Due to the extensive availability of license plate images from Belarus and Russia on Platesmania.com, we found it necessary to  limit the number of such license plates. This limitation was implemented to prevent a significant dataset imbalance and to ensure that the dataset benchmark and evaluation do not disproportionately favor these two countries. Specifically, we have constrained the number of images from Belarus and Russia to approximately 500,000 each. While we acknowledge that this still leads to imbalanced data, we anticipate that this undersampling strategy will contribute to a more accurate and globally representative benchmark for evaluating license plate recognition systems.

We divide the dataset into 60\% train, 20\% validation, and 20\% test. We also make sure not to have a similar issue as previous datasets, as described in \cite{near-duplicates}. We carefully check for matches within the splits using sequence analysis algorithms and ensure that license plates with the same/similar text are grouped under the same split. We set a threshold of 0.9 for Normalised Edit Distance, thus avoiding near duplicates from leaking into validation and test datasets. Normalised Edit Distance is defined as: $$\text{NED}(a, b) = \frac{\text{dist}(a, b)}{\max(len(a), len(b))}$$ where a and b are the predicted and ground truth recognition strings, dist is the Levenshtein distance, and len is the length of the strings a and b. Any text label with a 0.9 NED will be under the same split. 

As Platesmania images frequently feature only one prominent subject license plate per image, accompanied by a designated text recognition label, the splits for recognition (but not detection) exclusively incorporate images where the identification of the main license plate is unequivocal, particularly in scenarios involving multiple license plates within a single image. Official splits will be available at: https://github.com/siddagra/Global-License-Plate-Dataset

\begin{figure*}
    \centering
    \includegraphics[width=1\linewidth]{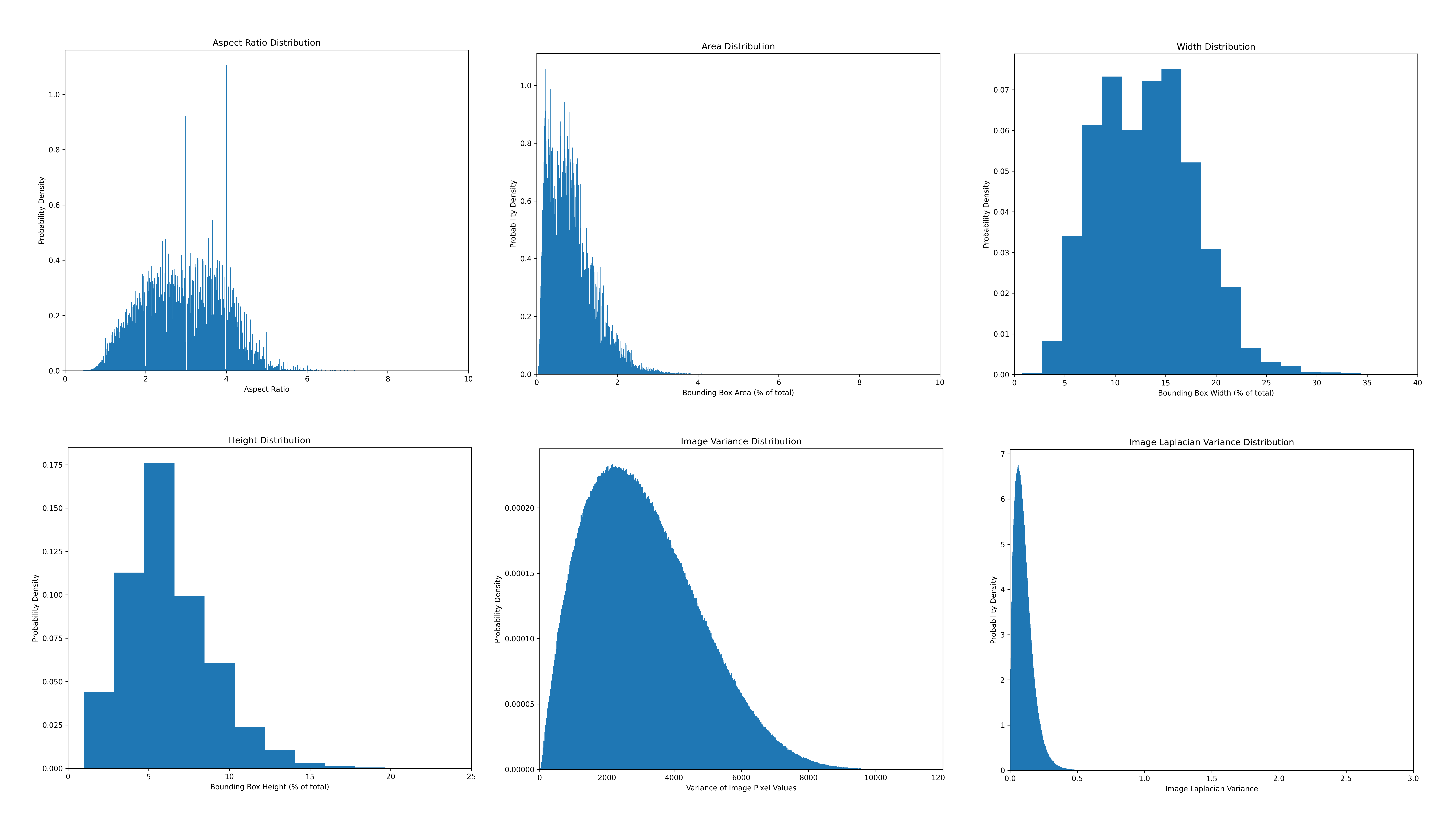}
    \caption{Relative Frequency Histograms of the Cropped License Plate Images}
    \label{fig:stats}
\end{figure*}

\subsection{Dataset Statistics}

As can be seen in \ref{fig:countr_hist} The dataset imbalance is quite high. In fact, Platesmania has even more images of Belarus and Russia, but they were limited in our split to allow for more balanced class representations.

As seen in \ref{fig:stats}, overall, while the dataset has variance in width and height, it also seems biased towards specific mean values. This is likely due to the fact that most images on Platesmania uploaded by users have the license plate and the vehicle as the central object of importance and therefore take pictures in similar mannerisms.

The variance of Laplacian was taken to quantify the 'edginess' or blurriness of the image. This was done via convolving the image with the Laplacian operator and computing its variance. Since the Laplacian operator is used for edge detection and localisation, it gives a rough approximation of the blurriness of the image. An Image Laplacian Variance of 0.5 suggests moderate level of detail and blurriness.

\subsection{Evaluation Metrics}
For license plate detection, the official metric should be mAP as popularised by many datasets such as Common Objects in Context (COCO) \cite{COCO}.

However, the primary metric of significance is the end-to-end license plate recognition. The official metric as popularised by \cite{xu2018towards} is accuracy, i.e., the number of license plates correctly identified divided by the total number of license plates. Even if a single character of the license plate is misrecognised, the entire label is considered incorrect.

\subsection{Model Overview}
License Plate Detection was trained on an NVIDIA RTX 3080 using YOLOv5 \cite{yolov5} model. The YOLOv5m model was trained for 130900 iterations with a batch size of 32, split evenly across 50 epochs, of which 3 epochs were for warm-up. Initial learning rate was set to 0.01 and highest learning rate was set to 0.1.

\begin{table}[ht]
\centering
\caption{YOLOv5 License Plate Detection Performance Metrics}
\begin{tabular}{lcc}
\hline
\textbf{Metric} & \textbf{Validation} & \textbf{Test} \\
\hline
mAP\textsuperscript{@0.50} & 95.38 & 94.85 \\
mAP\textsuperscript{@0.50:0.95} & 67.09 & 66.11 \\
F1\textsuperscript{@0.373} & 96 & 94 \\
Speed (ms) & 3.7ms  \\
Params & 21M  \\
\hline
\end{tabular}
\label{tab:metrics}
\end{table}
Note: F1-score was maximum (96\%) at a 0.373 confidence threshold.

\subsection{License Plate Recognition}

For the recognition model, as many License Plate Recognition models do not generalise to more characters, for example, \cite{scr_net}, \cite{zhang_wang_zhuang_2021} have fixed number of characters whereas we show baseline results by using PARSeq \cite{bautista_atienza_2022a} and CRNN \cite{CRNN}. Both were finetuned from pretrained weights as provided by \cite{bautista_atienza_2022a} for 33,600 iterations. Both models were designed with a maximum label length of 25 characters, suitable for capturing the variability in license plate text length across different jurisdictions. The batch size is set to 192 to balance computational efficiency and model performance, ensuring effective learning from diverse samples. Weight decay is deliberately set to zero to avoid unnecessary regularization effects, especially in such a large dataset. For CRNN, the learning rate is configured at 0.00051, incorporating a warm-up period using One Cycle Learning Rate Policy \cite{smith_topin_2017}. The warm-up constituted 7.5\%  of the total training steps to facilitate a more stable convergence. The input image size is set to 32x128 pixels, allowing the model to handle horizontal plate resolutions commonly encountered in real-world license plate images. The hidden size of the CRNN is set to 256, providing the model with sufficient capacity to capture complex patterns in the sequential data. PARSeq utilizes was set to a learning rate of 0.0007 with 7.5\% of training for warm-up. The patch size for PARSeq was set to 4x8 and embedding dimension was 384 with 12 attention heads. Dropout rate of 0.1 was applied to PARSeq. The last 25\% of the training steps were utilised for training the model using Stochastic Weight Averaging \cite{izmailov_podoprikhin_garipov_vetrov_wilson_2018}, with a learning rate of $1\times10^{-4}$, to improve generalisation. The dataset was trained on 60\% of the images and then validated on 20\% of the dataset. Finally, testing was done on 20\% of the remaining dataset. 

\begin{table}[h]
\centering
\caption{Model Accuracies}
\label{tab:model-accuracies}
\begin{tabular}{l c c c c}
\toprule
\textbf{Model} & \multicolumn{2}{c}{\textbf{Validation}} & \multicolumn{2}{c}{\textbf{Test}} \\
\cmidrule(lr){2-3} \cmidrule(lr){4-5}
 & \textbf{Accuracy\%} & \textbf{NED} & \textbf{Accuracy\%} & \textbf{NED} \\
\midrule
CRNN & 72.24 & 92.46 & - & -  \\
\textbf{PARSeq} & \textbf{91.88} & \textbf{97.01} & \textbf{87.93} & \textbf{92.24} \\
\bottomrule
\end{tabular}
\end{table}

Note: the reported accuracy is end to end accuracy. That is, the license plate recognition accuracy is reported with crops from license plate detection model (in this case YOLOv5m). The top confidence prediction from YOLOv5m was taken as input to the recognition model. Furthermore, accuracy is reported such that even if a single character is incorrectly predict in a license plate, the entire recognition is considered incorrect. However, Normalized Edit Distance (NED) corresponds to a more granular accuracy metric. Only PARSeq was tested on the test dataset to adhere to best practices involving holdout sets. This approach allows for a comprehensive evaluation of its performance and generalizability beyond the training and validation data, ensuring a robust assessment of the model's capabilities.

\section{Ethical Considerations and Limitations}
Our dataset compilation draws primarily from Platesmania.com, a platform generously providing us access to a vast repository of over 5 million license plates spanning 74 countries. Additional datasets, including CCPD, AOLP, Indian Commercial Truck License Plate Dataset, and Iraqi License Plate datasets, could be integrated in the future to enhance representation for under-sampled classes. Recognizing the inherent imbalance in class distribution, we took measures to address this bias by under-sampling majority classes. Specifically, the number of Belarus and Russia License Plates was limited to approximately 500,000 images each. However, despite these efforts, notable biases persist, with a significant imbalance in certain classes, as seen in \ref{fig:countr_hist}. The commitment to privacy and ethical data use extends to our 2023 version, which encompasses approximately 5 million images. To maintain a balanced dataset reflective of the research's initiation, we made a conscious decision to limit the number of additions. It is crucial to note that Platesmania continues to receive a constant influx of new images as users contribute fresh license plates regularly. As a result, our dataset remains dynamic, capturing the evolving landscape of license plate images available since the inception of our research.

To uphold the privacy and anonymity of individuals, our dataset includes best-practice measures such as blurring faces that may lead to additional identifiable information. In alignment with privacy considerations, diligent efforts were undertaken to anonymize and de-identify the dataset. Faces were systematically blurred to mitigate any potential risks associated with the inadvertent exposure of identifiable information. Watermarks were also blurred to the best of our abilities to reduce traces of identifiable information. These steps contribute to the anonymization and de-identification of the dataset, ensuring responsible use and adherence to ethical standards. 

Looking ahead, ethical considerations guide our ongoing efforts to expand the dataset responsibly. Platesmania's continual updates, introducing new license plate countries, inspire our commitment to incorporating these additions, promoting a more comprehensive and inclusive dataset. Moreover, we aspire to collaborate with other research papers and websites that house novel license plate datasets, particularly those representing under-represented categories. While this integration may introduce more images, ethical considerations remain at the forefront, emphasizing the importance of responsible data curation and usage.

In terms of community and stakeholder involvement, our dataset compilation process involved gathering images from Platesmania.com, a platform that engages a community of users perpetually uploading new license plates. While the dataset currently includes a curated subset collected until 2023, we recognize the dynamic nature of Platesmania, which continues to receive a substantial influx of images daily. We value ongoing community contributions and aim to regularly update our dataset to reflect the evolving landscape of license plate data.

Considering the potential algorithmic impact, we acknowledge the significance of under-sampled and biased classes on model performance. Efforts to address this involve intentional under-sampling of majority classes and continuous evaluation and refinement of the dataset. We are committed to conducting a thorough impact assessment, seeking to mitigate biases and improve the robustness and fairness of models trained on our dataset.

Looking ahead, our commitment extends to incorporating new license plate images from Platesmania and additional countries that have been added since the initiation of our research. Moreover, we aspire to collaborate with other researchers, websites, and publications that house small-scale novel license plate datasets, aiming to expand our benchmark and promote inclusivity and representation within the license plate recognition research community. We envision a collaborative and evolving dataset that addresses ethical considerations, encourages community engagement, and facilitates responsible algorithmic development.

\section{Conclusion}
We propose a diverse Global License Plate Dataset from across 74 countries, with various annotations such as license plate bounding box, 4-point corners, license plate segmentations, multi-class instance segmentations, car model, make, color, year, and license plate text labels. We achieve a baseline accuracy of 91.88\% on Validation dataset and 87.93\% on Test split using PARSeq on the dataset.

\bibliographystyle{plain}  
\bibliography{ref}  

\begin{thebibliography}{10}

\bibitem{media_lab}
Medialab lpr database, 2023.

\bibitem{Indian_Truck}
Siddharth Agrawal and Keyur~D. Joshi.
\newblock Indian commercial truck license plate detection and recognition for weighbridge automation.
\newblock In {\em 2022 28th International Conference on Mechatronics and Machine Vision in Practice (M2VIP)}, pages 1--6, 2022.

\bibitem{anagnostopoulos_anagnostopoulos_loumos_kayafas_2006a}
C.N.E. Anagnostopoulos, I.E. Anagnostopoulos, V.~Loumos, and E.~Kayafas.
\newblock A license plate-recognition algorithm for intelligent transportation system applications.
\newblock {\em IEEE Transactions on Intelligent Transportation Systems}, 7(3):377–392, Sep 2006.

\bibitem{baek_matsui_aizawa_2021}
Jeonghun Baek, Yusuke Matsui, and Kiyoharu Aizawa.
\newblock What if we only use real datasets for scene text recognition? toward scene text recognition with fewer labels.
\newblock {\em arXiv}, 2021.

\bibitem{bautista_atienza_2022a}
Darwin Bautista and Rowel Atienza.
\newblock Scene text recognition with permuted autoregressive sequence models.
\newblock {\em arXiv}, 2022.

\bibitem{bulan_kozitsky_ramesh_shreve_2017}
Orhan Bulan, Vladimir Kozitsky, Palghat Ramesh, and Matthew Shreve.
\newblock Segmentation- and annotation-free license plate recognition with deep localization and failure identification.
\newblock {\em IEEE Transactions on Intelligent Transportation Systems}, 18(9):2351–2363, Sep 2017.

\bibitem{goncalves_diniz_laroca_menotti_robson_schwartz}
Gabriel Gonçalves, Matheus Diniz, Rayson Laroca, David Menotti, and William Robson~Schwartz.
\newblock {\em Real-time Automatic License Plate Recognition Through Deep Multi-Task Networks}.

\bibitem{hsu2012application}
Gee-Sern Hsu, Jiun-Chang Chen, and Yu-Zu Chung.
\newblock Application-oriented license plate recognition.
\newblock {\em IEEE transactions on vehicular technology}, 2012.

\bibitem{hsu_chen_chung_2013a}
Gee-Sern Hsu, Jiun-Chang Chen, and Yu-Zu Chung.
\newblock Application-oriented license plate recognition.
\newblock {\em IEEE Transactions on Vehicular Technology}, 62(2):552–561, Feb 2013.

\bibitem{izmailov_podoprikhin_garipov_vetrov_wilson_2018}
Pavel Izmailov, Dmitrii Podoprikhin, Timur Garipov, Dmitry Vetrov, and Andrew~Gordon Wilson.
\newblock Averaging weights leads to wider optima and better generalization.
\newblock {\em arXiv}, 2018.

\bibitem{iraqi_license_plates}
nariman jabbar.
\newblock Dataset (iraq license plate), 2023.

\bibitem{deep_automatic_license_plate_recognition_system__proceedings_of_the_tenth_indian_conference_on_computer_vision}
Vishal Jain, Zitha Sasindran, Anoop Rajagopal, Soma Biswas, Harish~S Bharadwaj, and K~R Ramakrishnan.
\newblock Deep automatic license plate recognition system.
\newblock In {\em Proceedings of the Tenth Indian Conference on Computer Vision, Graphics and Image Processing}, ICVGIP '16, New York, NY, USA, 2016. Association for Computing Machinery.

\bibitem{yolov5}
Glenn Jocher, Alex Stoken, Jirka Borovec, NanoCode012, ChristopherSTAN, Liu Changyu, Laughing, tkianai, Adam Hogan, lorenzomammana, yxNONG, AlexWang1900, Laurentiu Diaconu, Marc, wanghaoyang0106, ml5ah, Doug, Francisco Ingham, Frederik, Guilhen, Hatovix, Jake Poznanski, Jiacong Fang, Lijun Yu, changyu98, Mingyu Wang, Naman Gupta, Osama Akhtar, PetrDvoracek, and Prashant Rai.
\newblock {ultralytics/yolov5: v3.1 - Bug Fixes and Performance Improvements}, October 2020.

\bibitem{near-duplicates}
Rayson Laroca, Valter Estevam, Alceu~S. Britto, Rodrigo Minetto, and David Menotti.
\newblock Do we train on test data? the impact of near-duplicates on license plate recognition.
\newblock In {\em 2023 International Joint Conference on Neural Networks (IJCNN)}. IEEE, June 2023.

\bibitem{li_shen_2016a}
Hui Li and Chunhua Shen.
\newblock Reading car license plates using deep convolutional neural networks and lstms.
\newblock {\em arXiv.org}, 2016.

\bibitem{li_wang_shen_zhang_2018}
Hui Li, Peng Wang, Chunhua Shen, and Guyu Zhang.
\newblock Show, attend and read: A simple and strong baseline for irregular text recognition.
\newblock {\em arXiv}, 2018.

\bibitem{COCO}
Tsung-Yi Lin, Michael Maire, Serge Belongie, Lubomir Bourdev, Ross Girshick, James Hays, Pietro Perona, Deva Ramanan, C.~Lawrence Zitnick, and Piotr Dollár.
\newblock Microsoft coco: Common objects in context, 2015.

\bibitem{loginov_2021}
Vladimir Loginov.
\newblock Why you should try the real data for the scene text recognition.
\newblock {\em arXiv}, 2021.

\bibitem{luo_jin_sun_2019}
Canjie Luo, Lianwen Jin, and Zenghui Sun.
\newblock Moran: A multi-object rectified attention network for scene text recognition.
\newblock {\em Pattern Recognition}, 90:109–118, Jun 2019.

\bibitem{narenbabu_sowmya_soman_2019}
R~Naren~Babu, V~Sowmya, and K~P Soman.
\newblock Indian car number plate recognition using deep learning.
\newblock {\em 2019 2nd International Conference on Intelligent Computing, Instrumentation and Control Technologies (ICICICT)}, Jul 2019.

\bibitem{confident_learning_cleanlab}
Curtis~G. Northcutt, Lu~Jiang, and Isaac~L. Chuang.
\newblock Confident learning: Estimating uncertainty in dataset labels, 2022.

\bibitem{CRNN}
Baoguang Shi, Xiang Bai, and Cong Yao.
\newblock An end-to-end trainable neural network for image-based sequence recognition and its application to scene text recognition, 2015.

\bibitem{shi_yang_wang_lyu_yao_bai_2019}
Baoguang Shi, Mingkun Yang, Xinggang Wang, Pengyuan Lyu, Cong Yao, and Xiang Bai.
\newblock Aster: An attentional scene text recognizer with flexible rectification.
\newblock {\em IEEE Transactions on Pattern Analysis and Machine Intelligence}, 41(9):2035–2048, Sep 2019.

\bibitem{smith_topin_2017}
Leslie~N Smith and Nicholay Topin.
\newblock Super-convergence: Very fast training of neural networks using large learning rates.
\newblock {\em arXiv}, 2017.

\bibitem{spanhel_sochor_juranek_herout_marsik_zemcik_2017}
Jakub Spanhel, Jakub Sochor, Roman Juranek, Adam Herout, Lukas Marsik, and Pavel Zemcik.
\newblock Holistic recognition of low quality license plates by cnn using track annotated data.
\newblock {\em 2017 14th IEEE International Conference on Advanced Video and Signal Based Surveillance (AVSS)}, Aug 2017.

\bibitem{tanwar_tiwari_chowdhry_2021}
Sanchit Tanwar, Ayush Tiwari, and Ritesh Chowdhry.
\newblock Indian licence plate dataset in the wild.
\newblock {\em arXiv}, 2021.

\bibitem{scr_net}
Yi~Wang, Zhen-Peng Bian, Yunhao Zhou, and Lap-Pui Chau.
\newblock Rethinking and designing a high-performing automatic license plate recognition approach.
\newblock {\em IEEE Transactions on Intelligent Transportation Systems}, 23(7):8868--8880, 2022.

\bibitem{efficient-teacher}
Bowen Xu, Mingtao Chen, Wenlong Guan, and Lulu Hu.
\newblock Efficient teacher: Semi-supervised object detection for yolov5, 2023.

\bibitem{xu2018towards}
Zhenbo Xu, Wei Yang, Ajin Meng, Nanxue Lu, and Huan Huang.
\newblock Towards end-to-end license plate detection and recognition: A large dataset and baseline.
\newblock In {\em Proceedings of the European Conference on Computer Vision (ECCV)}, pages 255--271, 2018.

\bibitem{xu_yang_meng_lu_huang_ying_huang_2018a}
Zhenbo Xu, Wei Yang, Ajin Meng, Nanxue Lu, Huan Huang, Changchun Ying, and Liusheng Huang.
\newblock Towards end-to-end license plate detection and recognition: A large dataset and baseline.
\newblock {\em Computer Vision – ECCV 2018}, page 261–277, 2018.

\bibitem{zhang_wang_zhuang_2021}
Yesheng Zhang, Zilei Wang, and Jiafan Zhuang.
\newblock Efficient license plate recognition via holistic position attention.
\newblock {\em Proceedings of the AAAI Conference on Artificial Intelligence}, 35(4):3438–3446, May 2021.

\end{thebibliography}

\end{document}